\documentclass{article}

\usepackage{microtype}
\usepackage{graphicx}
\usepackage{subfigure}
\usepackage{booktabs} % for professional tables
\usepackage{hyperref}

\usepackage{icml2018}

\begin{document}

\twocolumn[
\icmltitle{Exploring Optimal Control With Observations at a Cost}

% It is OKAY to include author information, even for blind
% submissions: the style file will automatically remove it for you
% unless you've provided the [accepted] option to the icml2018
% package.

% List of affiliations: The first argument should be a (short)
% identifier you will use later to specify author affiliations
% Academic affiliations should list Department, University, City, Region, Country
% Industry affiliations should list Company, City, Region, Country

% You can specify symbols, otherwise they are numbered in order.
% Ideally, you should not use this facility. Affiliations will be numbered
% in order of appearance and this is the preferred way.
\icmlsetsymbol{equal}{*}

\begin{icmlauthorlist}
\icmlauthor{Rui Aguiar}{equal,to}
\icmlauthor{Nikka Mofid}{equal,to}
\icmlauthor{Hyunji Alex Nam}{equal,to}
\end{icmlauthorlist}

\icmlaffiliation{to}{Stanford University, Stanford, CA, USA}
\icmlaffiliation{goo}{Stanford University, Stanford, CA, USA}
\icmlaffiliation{ed}{Stanford University, Stanford, CA, USA}

\icmlcorrespondingauthor{Cieua Vvvvv}{c.vvvvv@googol.com}
\icmlcorrespondingauthor{Eee Pppp}{ep@eden.co.uk}

% You may provide any keywords that you
% find helpful for describing your paper; these are used to populate
% the ``keywords`` metadata in the PDF but will not be shown in the document
\icmlkeywords{Machine Learning, ICML}

\vskip 0.3in
]

% this must go after the closing bracket ] following \twocolumn[ ...

% This command actually creates the footnote in the first column
% listing the affiliations and the copyright notice.
% The command takes one argument, which is text to display at the start of the footnote.
% The \icmlEqualContribution command is standard text for equal contribution.
% Remove it (just {}) if you do not need this facility.

%\printAffiliationsAndNotice{}  % leave blank if no need to mention equal contribution
% \printAffiliationsAndNotice{\icmlEqualContribution} % otherwise use the standard text.

\begin{abstract}
%This document provides a basic paper template and submission guidelines.
%Abstracts must be a single paragraph, ideally between 4--6 sentences long.
%Gross violations will trigger corrections at the camera-ready phase.
There has been a current trend in reinforcement learning for healthcare literature, where in order to prepare clinical datasets, researchers will carry forward the last results of the non-administered test known as the last-observation-carried-forward (LOCF) value to fill in gaps, assuming that it is still an accurate indicator of the patient's current state. These values are carried forward without maintaining information about exactly how these values were imputed, leading to ambiguity. Our approach models this problem using OpenAI Gym's Mountain Car and aims to address when to observe the patient's physiological state and partly how to intervene, as we have assumed we can only act after following an observation. So far, we have found that for a last-observation-carried-forward implementation of the state space, augmenting the state with counters for each state variable tracking the time since last observation was made, improves the predictive performance of an agent, supporting the notion of ``informative missingness", and using a neural network based Dynamics Model to predict the most probable next state value of non-observed state variables instead of carrying forward the last observed value through LOCF further improves the agent's performance, leading to faster convergence and reduced variance.
\end{abstract}

\section{Motivation}
\label{submission}

%// talk about the problem and the motivating example
%Submission to ICML 2018 will be entirely electronic, via a web site
%(not email). Information about the submission process and \LaTeX\ templates
%are available on the conference web site at: The guidelines below will be enforced
%for initial submissions and
%camera-ready copies. Here is a brief summary:
In the medical setting, a clinician often decides between administering two costly tests that give some underlying information about the patient's condition \cite{Fleming2019MissingnessAS} \cite{Buck1960AMO}. If one test is chosen to be administered, it is currently common practice in hospitals to update the timestamp of the test's administration, leaving that of the other test the same. Because of this practice, however, raw clinical data is not necessarily in a form that RL agents can easily learn from. Thus, there has been a current trend in reinforcement learning for healthcare literature, where in order to prepare clinical datasets, researchers will carry forward the last results of the non-administered test known as the last-observation-carried-forward (LOCF) values to fill in gaps, assuming that they are still an accurate indicator of the patient's current state. These variables are carried forward without maintaining information about how their values were imputed, leading to ambiguity. As RL becomes more ubiquitous, it is the high level goal to create an agent that can help improve patient outcomes, which implicitly requires knowing when to observe the patient's physiological state (using any one of a number of available laboratory tests) as well as when/how to intervene. Our method, detailed further in the following section, aims to address when to observe the patient's physiological state and how to act optimally both in terms of saving the cost of unnecessary tests and accurately maintaining the patient's health status. 

\section{Approach}
\label{submission}
%// talk about the problem and the motivating example
%Submission to ICML 2018 will be entirely electronic, via a web site
%(not email). Information about the submission process and \LaTeX\ templates
%are available on the conference web site at: The guidelines below will be enforced
%for initial submissions and
%camera-ready copies. Here is a brief summary:
We are modeling the issue of optimal control with observation cost using the Open AI Gym Mountain Car simulator. The simulator gives an agent two pieces of information about the game state: the car position and velocity. The game objective is to get the car to the goal position defined by the y position of the car in as few steps as possible. In the original setup, the agent only has three actions: to push the car left, right, or apply no additional force. In order to model the medical setting, we assume that one can observe the car's position or velocity but at a significant negative cost. In turn, we adjust the state and action space, making each state a tuple of the car's current position and velocity, carrying forward the last observation of the value that was not observed, henceforth referred to as the ``LOCF without Counters" state space, and expanding the car's action space so each action becomes two-fold, both moving the car and deciding whether to observe one, both, or neither of the variables. Next, we augment this initial state space representation, adding a counter for both position and velocity in order to track the number of timesteps since their last observation, henceforth referred to as the ``LOCF with Counters" state space. Finally, we implement a multi-layer perceptron forward dynamics model, on top of our current DQN implementation, to predict the most probable next state value of any non-observed state variables, providing a better approximation of non-observed values than the outdated values carried over from previous time steps.

We have successfully been able to implement our augmented action space, reward space, and proposed state spaces,``LOCF without Counters" and ``LOCF with Counters", and have been able to solve mountain car with these state spaces using a DQN variant with a mechanical energy based reward, as well as implemented our dynamics model to predict the next position and velocity of the car given the current state and the agent's chosen action. As we will detail in this report,     we have discovered that the addition of counters keeping track of the age of an observation to the state space led the agent to learn to reach the goal state more rapidly than without, with significantly low variance. Furthermore, the addition of a dynamics model was able to further improve our agent's performance showing that through the use of a neural network trained on the rollout data, we can find a better way to impute the values of the missing variables than the current practice of LOCF.

\section{Related Work}
%Write up our related work
Our work is based heavily on concepts drawn from the paper: ``Missingness as Stability: Understanding the Structure of Missingness in Longitudinal EHR Data and its Impact on Reinforcement Learning in Healthcare" written by Scott Fleming. In this paper, Fleming explains the concept of ``informative missingness", which is the notion that missingness in data can provide invaluable information about the patient state since the clinician must have considered a variety of underlying factors in deciding whether or not to administer a test. It is important to note that most of these factors are not necessarily captured in the medical documents but the existence of such factors may still be inferred through missingness. \cite{Fleming2019MissingnessAS} \cite{Sharafoddini2019ANI} \cite{Che2018RecurrentNN}. Specifically, the paper shows that incorporating missingness into the state representation, as a binary indicator, improves the prediction of patient state dynamic \cite{Fleming2019MissingnessAS}. In this paper, we are expanding on the ideas in Fleming's paper and exploring more complex indicators, in this case keeping track of the age of a carried forward observation through counters, to see how they affect our agent's performance and then building on top of that using a neural network to estimate the missing states instead of re-using outdated values.

\section{Augmented Environment Details}
\label{submission}
In addition to the basic setup explained in Approach, our Mountain Car has the following state space augmentation to account for the healthcare analogy:

\subsection{State Space Augmentation:}
We created two main state space representations: ``LOCF without Counters" and ``LOCF with Counters". Both representations include the car's velocity and position. If the agent chooses to observe one or both of the variables, the according value of position or velocity will be updated with current information about the car. If a value is not observed, the last obserervation of the variable will be carried forward and assumed to be an accurate representation of the car's current state, as is currently the precedent by RL researchers in preparing clinical datasets. Additionally, ``LOCF with Counters" includes counters for both position and velocity tracking how many time steps have elapsed since the last time each of the values were observed, adding information to the state space about the age of an observation. We eventually expand upon our ``with Counter" state space by replacing the LOCF -- or the carrying forward of the last observation of a non-observed state variable -- with the most probable next state value predicted from a neural network based Dynamics model of the car. We will refer to this variation of the ``with Counter" state space as "Dynamics Model with Counters."

\subsection{Action Space Augmentation:} 
We expanded the action space as follows: [0, ..., 11] where each of the twelve values corresponds to the two-fold action of the car moving left, right, or none, and deciding whether to observe either or both of the position and the velocity. 

\subsection{Reward Space Augmentation:} 
In our mountain car environment, -1 is rewarded for each time step spent until the agent reaches the goal height of 0.5. An additional cost of -8 is incurred for observing either the position or the velocity. Initially, we added the negative absolute difference between the car's current position and the goal position as an additional reward term to encourage the agent's learning. Ultimately, we decided to use the DQN with mechanical energy based reward, without the negative absolute distance reward. 

\section{Algorithms}

We implemented the following algorithms and will discuss the result from each one below:

\subsection{SARSA with Function Approximation}
The first algorithm we explored to solve the mountain car problem was SARSA with Function Approximation \cite{SamKirkiles}. We were able to get this algorithm to solve vanilla mountain car relatively quickly, but once we augmented the action and state space for ``LOCF with Counters" and ``LOCF without Counters"  it was no longer able to solve the problem. We noticed that it would not reach the goal state with 20,000 steps per episode (while for vanilla it could reach the goal state in less than 200 steps). We hypothesized that this was because of the augmented state and action space, which significantly increased the complexity of the task, compared to the vanilla mountain car problem.

\subsection{Q learning with Function Approximation}
After trying SARSA with Function Approximation, we also implemented Q-learning with Function Approximation to try and solve both vanilla mountain car and mountain car with our ``LOCF with Counters" and ``LOCF without Counters" state spaces. Similar to SARSA with function approximation, Q-learning with function approximation was able to solve vanilla mountain car relatively quickly.  However, it began to struggle when we introduced the LOCF implementation of the problem. For both the LOCF state spaces, Q-learning failed to solve mountain car after training for hundreds of iterations.  

\subsection{DQN Variant with Mechanical Energy based Reward}
Because both SARSA and Q-learning failed to learn mountain car with the last-observation-carried-forward (LOCF) condition both with and without counters, we decided to adopt a more robust, reward-shaping algorithm. We used a DQN variant employing a special reward function derived from the mechanical energy of the car to provide the agent with incentives based on the car's dynamics, code written by Sachin \cite{Sachin1} . The reward is as follows.

$100*((sin(3*p_{t+1}) * 0.0025 + 0.5 * v_{t+1} * v_{t+1}) - (sin(3*p_t) * 0.0025 + 0.5 * v_t * v_t))$

In the above reward scheme, ($p_t, v_t$) is the position and velocity at time step $t$ and ($p_{t+1}, v_{t+1}$) is the position and velocity at the next time step. We also subtracted our observation cost of -8 from this reward when we took an observation action\footnote{Note that the observation cost of 8 is determined through some level of hyper parameter search. We've also experimented with the observation cost of -3 and -2 to reduce the negative incentives associated with making an observation, but based on the learning curves concluded that the observation cost of -8 is representative of any environments with high observation cost.}. This approach was able to solve the mountain car problem for both the vanilla, ``LOCF without Counters", and  ``LOCF without Counters" state space representation, with the ``LOCF with Counters" performing significantly better than the ``LOCF without Counters" state space with low variance improvement, demonstrating that adding the counters makes the problem more tractable to solve despite the high cost of observation.

\subsection{Multi Layer Perception For Forward Dynamics Model}
After training the DQN variant with the Mechanical Energy Based reward, which was able to solve mountain car for both the ``LOCF with Counter" and the ``LOCF without Counters" state space, we decided to train a neural network to estimate the physics of the mountain car problem -- acting as a dynamics model, which we could use to predict the most probable next state value of a non-observed state variable instead of carrying forward its last observed value with LOCF. We used a multi-layer perceptron feed forward neural network as our model architecture and we trained it offline with data gathered by running our DQN with Mechanical Energy Reward for the ``LOCF with Counters" state space and recording the state, action, reward, next state, and next action (s,a , r, s',a') tuples. Once the model was trained, we ran it using the ``with counters" state space augmentation replacing the carrying forward of the last observation of a non-observed state variable (``LOCF'), with a prediction from our dynamics model of the most probable next state value to be used as the representation of the current non-observed state value instead. Specifically, we replaced the carried forward position and velocity with forward pass through the model to get an estimate for the next state, given the current position, velocity, and action. This change further reduced the difficulty of the problem while also enhancing the stability of convergence and the performance of our agent, leading it to arrive at the goal position much more quickly than the one using last-observation-carried-forward with counters. The forward dynamics modelling can be applied to any medical settings where state variables are correlated (e.g. aPTT and anti-Xa are correlated tests that inform a clinician of an appropriate anticoagulant dosage for an ICU patient though in real medical dataset, the correlation may be obscured by noise) \cite{Fleming2019MissingnessAS}.

\section{Experiment Details}
%Explain the number of episodes, steps, time we trained for
We trained our DQN variant with Mechanical Energy based Reward with the different state space augmentations described ``LOCF without Counters" and ``LOCF with Counters", as well as extending our ``with Counter" state space to use a dynamics model of the car to predict the most probable next state values for non-observed state variables instead of last-observation-carried-forward, referred to as ``Dynamics Model with Counter." For all our experiments, we trained our DQN for 1,000 episodes, capping the number of steps at 20,000. We used a learning rate of 0.001 and an initial epsilon of 1 which we annealed by a decay of 0.995 every iteration. We also have a discount factor or ``gamma" of 0.95 and the replay buffer of our DQN also uses a batch size of 64. It took approximately 8 hours to train our DQN without the use of counters and 5 hours to train our DQN with the use of counters in the state space.

\section{Experiment Results}
The following plots show the learning curves for the different state space augmentations paired with our DQN with Mechanical Energy Based Reward: ``LOCF without Counters", ``LOCF with Counters," and ``Dynamics Model and Counters." Looking at the plots, it is easy to see that the addition of counters as well as the eventual replacement of the LOCF for missing states with the predictions of the dynamics model leads to faster convergence and minimized variance as compared to the ``without Counter" state space.

    \begin{figure}[h!]
  \caption{DQN for 1K training eps --  ``LOCF without Counters"}
  \centering
    \includegraphics[width=7cm]{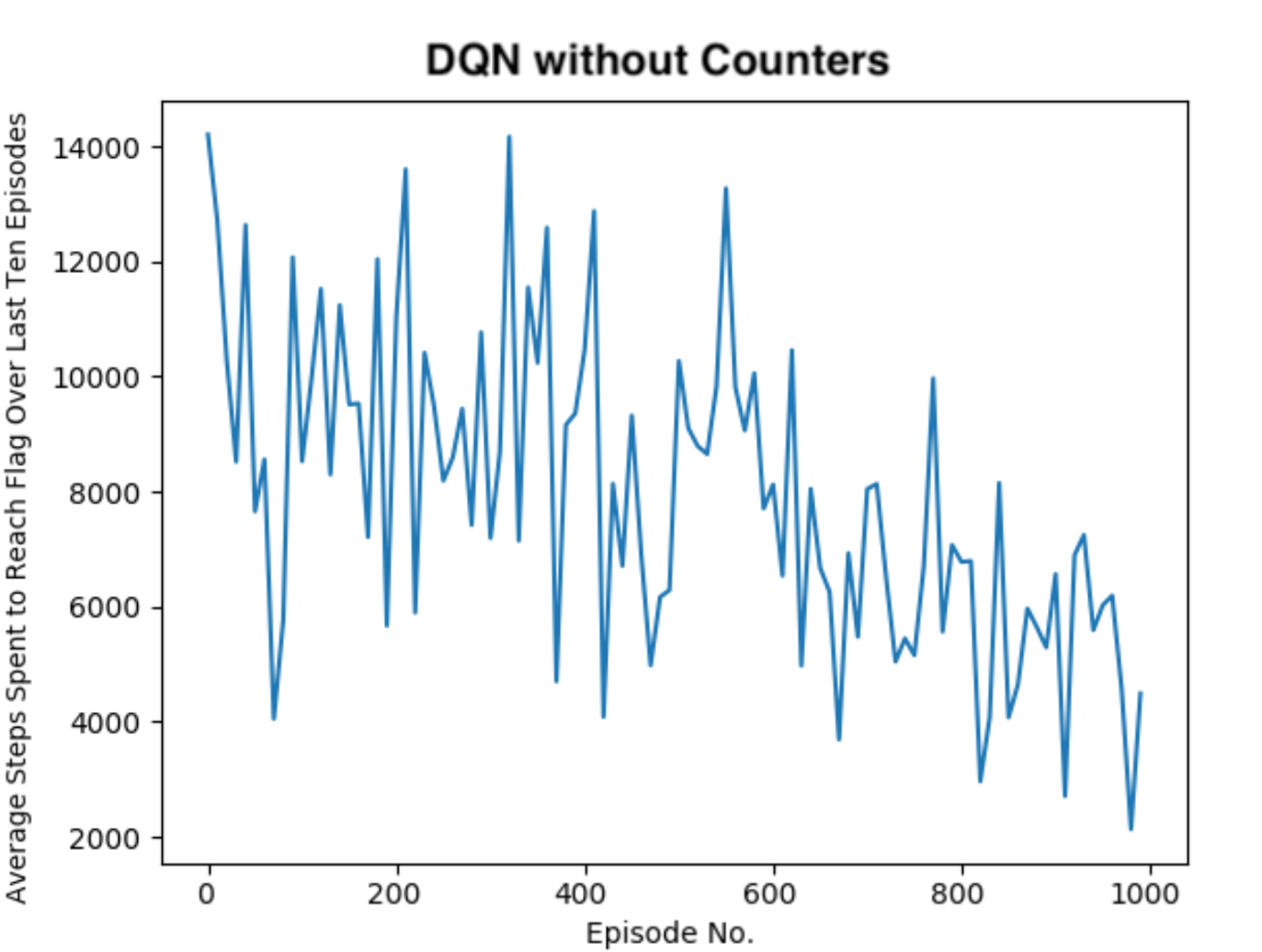}
    \end{figure}

\begin{figure}[h!]
  \caption{DQN for 1K training eps -- ``LOCF with Counters"}
  \centering
    \includegraphics[width=7cm]{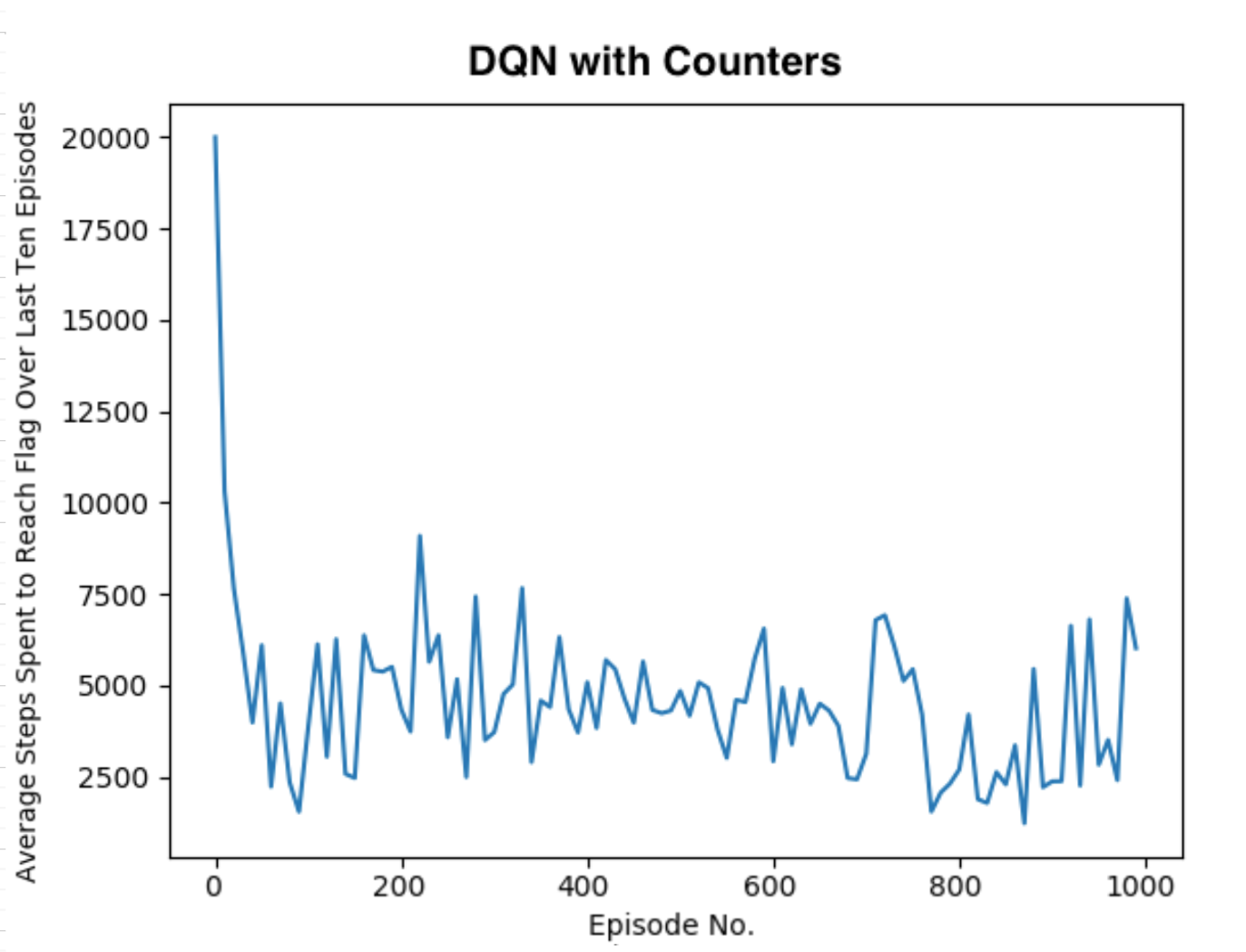}
    \end{figure}

 \begin{figure}[h!]
  \caption{DQN for 1K training eps -- ``Dynamics Model with Counters"}
  \centering
    \includegraphics[width=7cm]{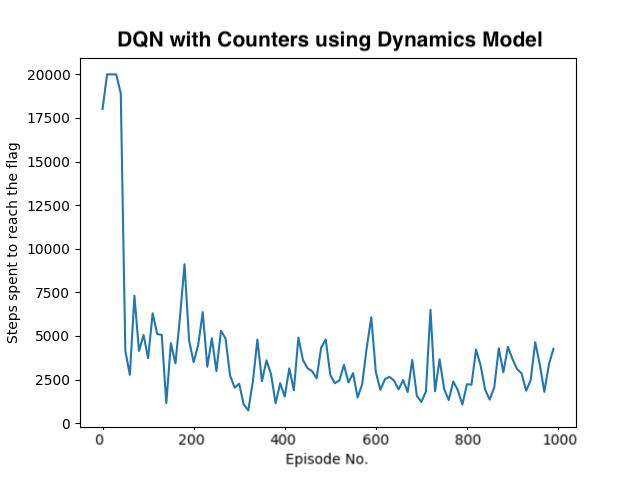}
    \end{figure}

 \begin{figure}[h!]
  \caption{Comparison of All Three Methods}
  \centering
    \includegraphics[width=7cm]{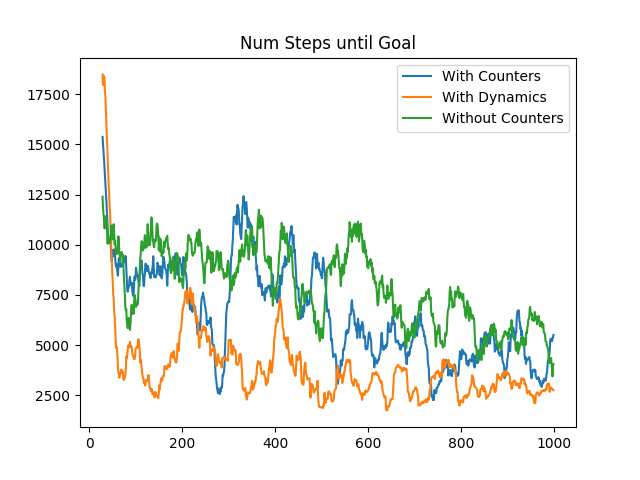}
    \end{figure}

\section{Qualitative Results}
In order to provide more qualitative analysis, we generated the following action histograms which show the percentage of observation of position and velocity based on the car's $x$ position. We also include plots that give information about the ``Ratio of Position and Velocity Observation" as well as ``No Observations to the Total Actions" for each episode in the Appendix in order to give information about the car's observation behavior over time with the different state spaces. From these plots, we can gain insight into the behavior of the car and the effect of the different state space augmentations on the agent.

    \begin{figure}[h!]
  \caption{Percent of Position and Velocity Observation Based on Car Position Over 1k Episodes: ``LOCF without Counters"}
  \centering
    \includegraphics[width=7cm]{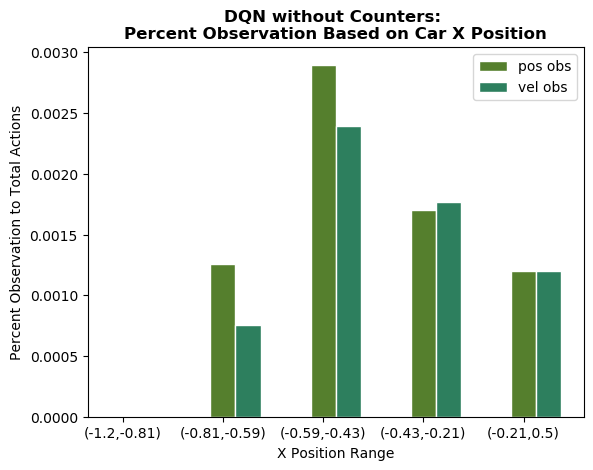}
    \end{figure}

\begin{figure}[h!]
  \caption{Percent of Position and Velocity Observation Based on Car Position Over 1k Episodes: ``LOCF with Counters"}
  \centering
    \includegraphics[width=7cm]{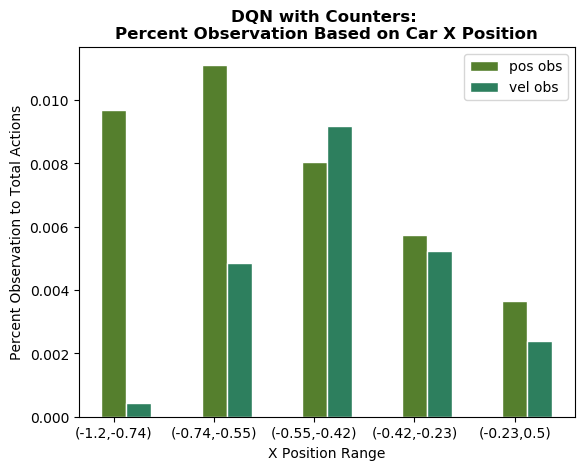}
    \end{figure}

 \begin{figure}[h!]
  \caption{Percent of Position and Velocity Observation Based on Car Position Over 1k Episodes: ``Dynamics Model with Counters"}
  \centering
    \includegraphics[width=7cm]{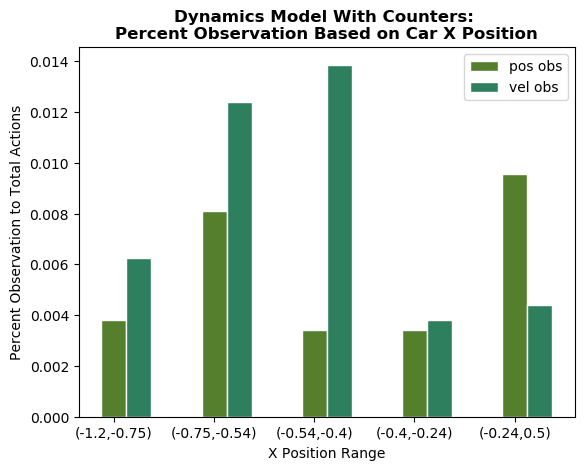}
    \end{figure}

\section{Analysis and Discussion}
Our analysis is two-fold: First, we looked into how the state augmentation with counter values enhanced the agent's performance; next, we observed action distribution across different positions of the car and the number of training episodes to analyze what has the agent learned to act optimally -- when and how the agent is choosing to observe in order to minimize the cost while still reaching the goal. In addition to generating learning curves that show the number of steps per episode to reach the goal as shown in Figures (1)-(3), we examined the action histograms, highlighting the frequency of observing versus non-observing actions, dependent on the car's position as shown in Figures (4)-(6) and on the number of training episodes (review Appendix). 

\subsection{Experimental Results Analysis}

First, we examined the agent's performance with and without state augmentation to verify the notion of informative missingness. Based on the experiment results shown in Figures (1), (2), and (4) for the ``LOCF without Counters" and `` LOCF with Counters" state space, we can see that the agent supported by the augmented state representation clearly outperforms the agent without the state space augmentation. When aided by the counters, which provide information about the age of velocity and position, the agent is able to reach the goal much more quickly as shown in Figure (2) and with reduced variance. In contrast, when the agent is only given the position and velocity observations without any information about the number of timesteps these variables have been carried over for, it takes over 200 episodes for the agent to reach the goal in less than 10,000 steps. Although the agent eventually learns to achieve the goal in less than 4,000 steps in both the augmented and non-augmented state spaces, the agent does not experience as steep of a learning curve as the agent with the counter information. Our result is consistent with the claim about informative missingness made in Fleming et al.'s paper -- providing the age of information, and whether or not certain pieces of medical information have been updated, can provide invaluable insights into the patient's condition, significantly enhancing the RL agent's predictive performance of the patient's status \cite{Fleming2019MissingnessAS}. 

Finally, from Figure (3) we see that the replacement of the last-observation-carried-forward with the prediction of the most probable next state value for a non-observed state variable through the use of the neural network Dynamics model further leads to improved predictive performance and reduced variance. This is further backed by Figure (4) which shows that the use of ``Dynamics Model and Counters" performs better than ``LOCF without Counters" and ``LOCF with Counters". This supports our initial hypothesis that using a neural network to predict the next state value instead of carrying forward the value of the last observation for a non-observed state variable could be a viable and powerful method to improve RL datasets, particularly when combined with counters to keep track of number of timesteps since the last observation an shows that utilizing such a Dynamics model along enhancing a state space with counters tracking the age of information could be a great method to combat missingness in RL datasets. Thus through our experimentation with ``LOCF without Counters", ``LOCF with Counters", and ``Dynamics Model with Counters", we have supported the notion of informative missingness and enhanced the predictive performance of our agent.

\subsection{Qualitative Results Analysis}
%\#\#(This first paragraph could be combined with 9.1? into a section about state augmentation with counters..?)

%An analysis of the convergence in figures (1) through (3) displays the importance of informative missingness and the dynamics model we trained. In figure (1) without counters, the DQN is very noisy and takes a long time to converge However, when we add the counters, we suddenly see a sharp convergence over the first few iterations, followed by a slightly noisy result where our model converges around 3000 iterations. Finally, when we use the counters in figure (3), we see a similar behavior - the model sharply converges after the first few iterations, then continues to lower until it converges to a noisy range around 2500 iterations. 

In order to better understand the behavior of our car, we generated a number of histogram plots which show the ``Percent Observation Based on the Car $x$ Position." The histogram plots for the models that we ran illuminate how the model is learning across iterations, and specifically how changing the state space representation influences when and where the mountain car agent takes observation as it attempts to scale the cliff to reach the goals state of the flag. Before we analyze the histograms, it is important to note that the mountain car position ranges from -1.2 to 0.6 inclusive with 0.5 marking the goal state. We split the range of positions that the car actually reaches into 5 smaller brackets of position for ease of analysis which explains why the position 0.5 is the maximum value of the the 5th bucket in our histograms, as the car reaches the goal state at $x = 0.5$, but doesn't go past it to 0.6.

Figure (5) shows the ``Percent of Position and Velocity Observation Based on Car Position" for over 1,000 episodes for the ``LOCF without Counters" state space. In Figure (5), we can see that position and velocity observations are approximately equal, but interestingly, there are no observations in the range (-1.2, -.81). This may be because the car gets stuck in a greedy loop trying to go up the hill to gain a small amount of mechanical energy reward, and reaches very few if any states  on the far left of the slope which consistent with the ``without Counter" model taking a long time to converge, as it is very infrequently going to the back of the slope, a necessary requirement for completing mountain car. 

Figure (6) which analyzes the car's observations across X positions as it attempts to reach the goal state for the ``LOCF with Counters" state space, we can see by looking at the scale of the y axis that the model chooses to observe far more than without counters. This is interesting because the better performance of the model, coupled with the increased observation rate implies that the model is learning when to observe, and using the observations to solve mountain car more quickly. Additionally, the DQN seems to be observing far less in the later buckets when it is closer to the goal state - implying that as we get closer to the goal, the car is able to realize that it is in a favorable position and accelerate towards the finish without making more costly observations. 

Finally, Figure (7) showcases the ``Percent of Position and Velocity Observations" based on Car Position over 1,000 episodes for "Dynamics Model with Counters." At first glance, it appears that this state space representation has lead to increased velocity observations, however, if we take a closer look at Figure (9) in the appendix which shows the ``Ratio of Position and Velocity Observations to Total Actions" for each episode for the ``Dynamics Model with Counters" representation, we can see that almost all of these observations came in the first few episodes, when we were almost completely exploring instead of exploiting. Therefore, if you discount the velocity observations you can see that the model is actually observing far less than in figure (6) - which is what you would expect if the agent does not need to observe any of the variables because the dynamics model predicts approximately correct position and velocity values based on the previous position and velocity at a zero cost. This implies that the DQN is learning that dynamics model is providing a "pseudo observation" sufficiently accurate to base its actions on, thus optimizing to observe far less, especially because observation comes at such a high cost and provides little help.

In summary, adding both counters and the dynamics model improves the performance of our mountain car DQN agent. Without these two variants, we suspect that LOCF is a hard problem for the DQN to solve due to the missing variables. Additionally, if we look at the distribution of the observations our model makes along the $x$-axis, we can see that the agent learns to observe more to complete its goal when the counters are introduced, and observe less again when the dynamics model is introduced. This is consistent with our understanding of informative missingness because the results indicate that the age of information can signal to an agent about when to observe. When the forward dynamics predictor is introduced, the agent no longer needs to rely on expensive observations since the approximate values can be obtained at no cost and therefore learns to reach the goal without making observation. It is also interesting to observe that in both ``LOCF Without Counters" and values imputed by forward dynamics, the agent avoids making expensive observations, but in the first case the negative incentive associated with observation impedes the agent's learning while in the latter case, the observation cost does not affect the agent's learning. 

\subsection{Statistical Analysis}
Next, we were interested in statistically examining the correlation between the $x$ position of the car and whether or not the agent makes observation. Based on the data collected from experiments with and without counters as well as the experiment with the forward dynamics predictor, we ran logistic regression with $x$ position of the car as the independent variable and whether or not the agent observed one of the game variables as the dependent variable, and observed the p-value of the $x$ coefficient. With `LOCF No Counters' data, $x$ position of the car and whether or not the agent observed velocity has the p-value of $0.004$. Similarly, with `Values Imputed by Forward Dynamics' data, the p-value of the two variables is $0.003$. With `LOCF with Counters', the p-value is $0.041$, which is higher than the first two cases but still statistically significant. In all three cases, the p-value is statistically significant to reject the null hypothesis that the agent's velocity observation is independent of the car's current position along the $x$-axis of the slope. We were also interested in observing a similar trend with the agent's position observation, but since the number of position observing samples is too small, compared to the total number of actions, in most of our datasets, we ran into perfect or quasi separation issue when training the logistic classifier. While we were unable to successfully reject the null hypothesis for the agent's observing position, the p-testing on the agent's velocity observation suggests a strong correlation between the car's position and whether or not to observe velocity. 

\section{Conclusion and Next Steps}
Through this paper, we hoped to (1) better establish the notion of informative missingness with our mountain car state augmentation, (2) explore an alternative to LOCF and improve the agent's predictive performance directly through a forward dynamics model to compensate for the missing observations, and (3) examine what leads to the agent's choosing to observe or not. We've successfully answered the first two questions regarding ways to improve the agent's performance in the face of missing data due to costly observations. We are also interested in different ways to verify the factors leading to the agent's decision to observe or not. Our current hypothesis involves the car's position along the x-axis and the number of training episodes, but this hypothesis testing would require more training samples to reduce variance in our results and be able to identify the correlation between the agent's actions and different environment variables. 

Our current next steps are: (1) Reduce variance and further verify reproducibility of the experiments by running the same experiment with different seeds. 
(2) Apply to medical simulators where correlations between state, action, and next state are less conspicuous from the dataset. Again test the notion of informative missingness with a simple state augmentation and train a predictor to identify the correlation, then test the effectiveness of the dynamics predictor. 
(3) Further complicate the mountain car example by training the forward dynamics online. In our current experiment setup, the dynamics model is trained on the dataset from a previous rollout. Instead, we propose to train the dynamics model as the agent is learning to observe or not observe the game variables. In this setting, the observation has an additional benefit of providing the dynamics model with reliable data from which the correlation can be learned. It would be interesting to observe whether the agent becomes aware of this additional advantage when choosing to observe. In such a case, we predict that the agent would observe both the velocity and the position significantly more in earlier training episodes to help train a more reliable dynamics predictor, then in later episodes, would optimize towards not making any more observation and simply obtain estimated values from the predictor. This last extension would imply that expensive observations may be more useful in the earlier stages of learning when the agent is trying to learn the underlying correlation of the state and the next state given some action, and as the dynamics model becomes more reliable, the agent relies less on making expensive observation.

\newpage

% In the unusual situation where you want a paper to appear in the
% references without citing it in the main text, use \nocite
\nocite{langley00}

\bibliography{example_paper}
\bibliographystyle{icml2018}

\section{Appendix}
Review the next page for plots referenced throughout the paper.
\newpage
\onecolumn
\begin{figure}[h!]
\hfill
\subfigure[Ratio of Position Observations to Total Action]{\includegraphics[width=5cm]{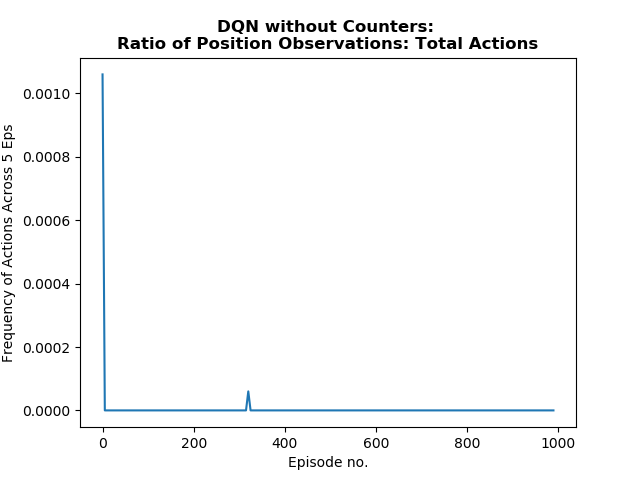}}
\hfill
\subfigure[Ratio of Velocity Observations to Total Action]{\includegraphics[width=5cm]{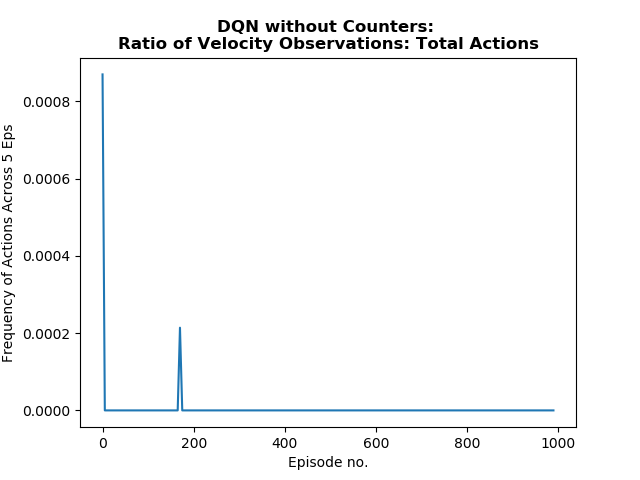}}
\hfill
\subfigure[Ratio of No Observations to Total Action]{\includegraphics[width=5cm]{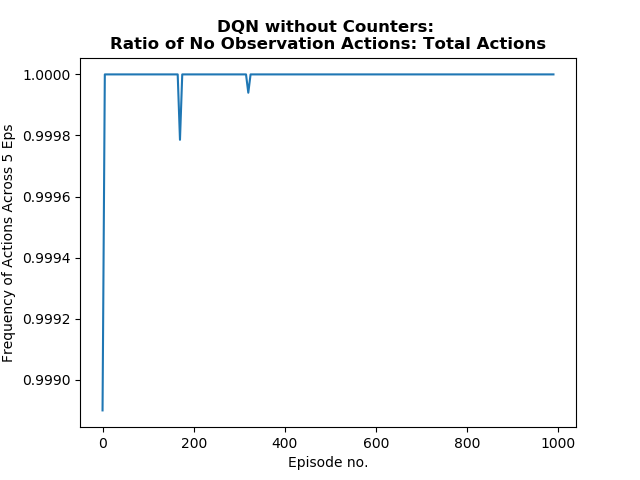}}
\hfill
\caption{Ratio of Observation to Total Action for "LOCF without Counters"}
\end{figure}

\begin{figure}[h!]
\hfill
\subfigure[Ratio of Position Observations to Total Action]{\includegraphics[width=5cm]{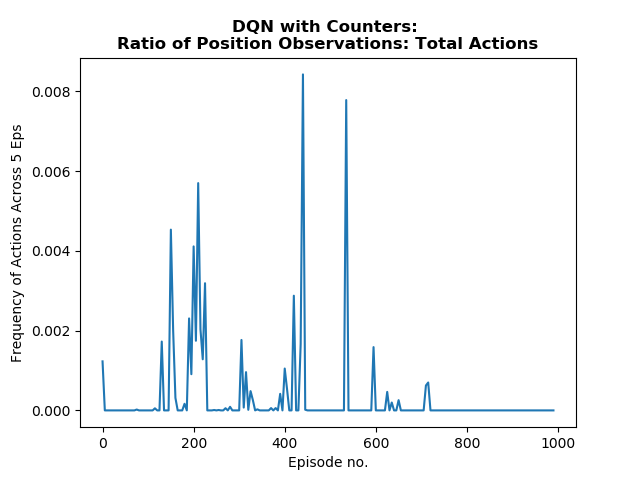}}
\hfill
\subfigure[Ratio of Velocity Observations to Total Action]{\includegraphics[width=5cm]{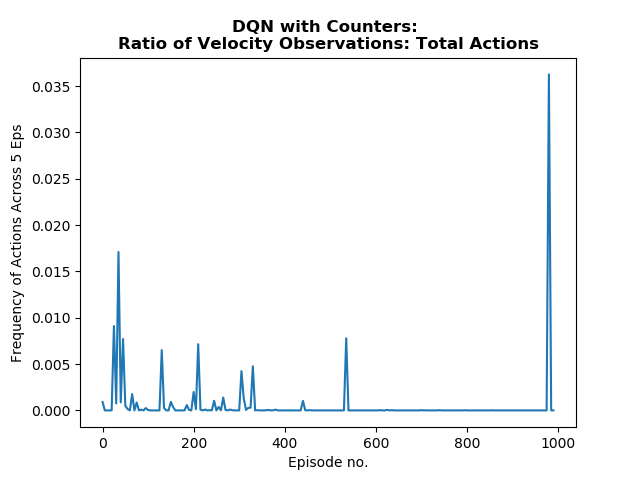}}
\hfill
\subfigure[Ratio of No Observations to Total Action]{\includegraphics[width=5cm]{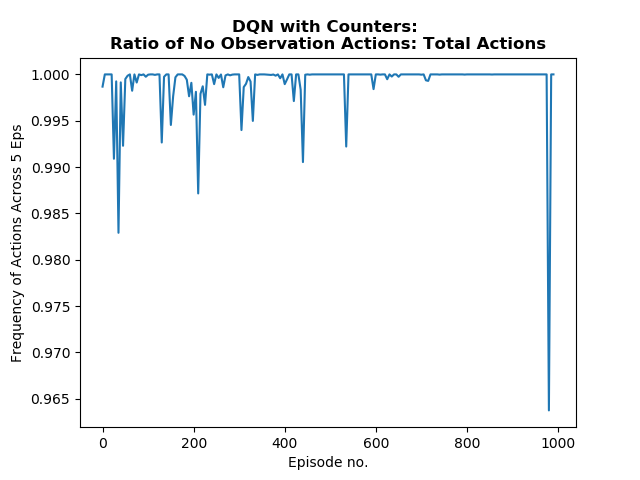}}
\hfill
\caption{Ratio of Observation to Total Action for "LOCF with Counters"}
\end{figure}

\begin{figure}[h!]
\hfill
\subfigure[Ratio of Position Observations to Total Action]{\includegraphics[width=5cm]{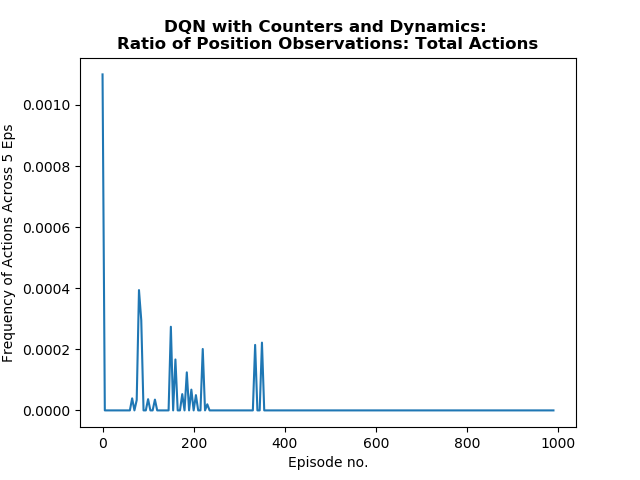}}
\hfill
\subfigure[Ratio of Velocity Observations to Total Action]{\includegraphics[width=5cm]{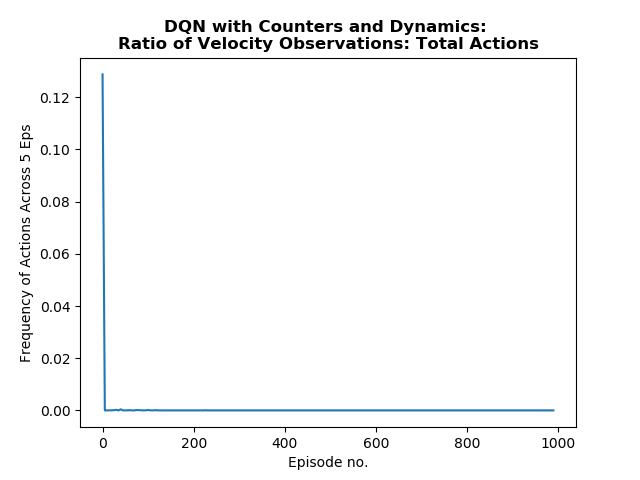}}
\hfill
\subfigure[Ratio of No Observations to Total Action]{\includegraphics[width=5cm]{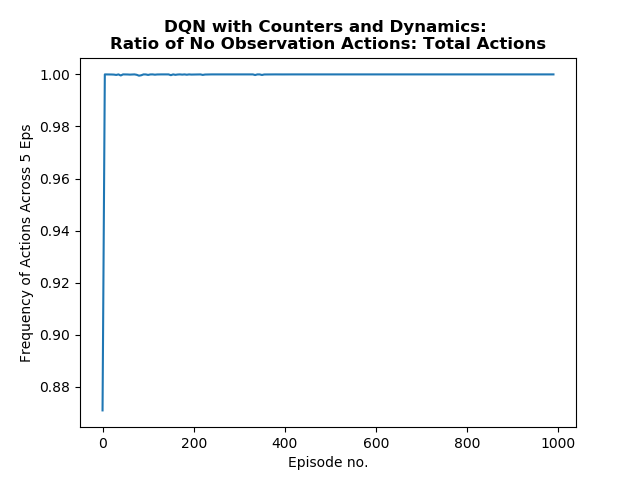}}
\hfill
\caption{Ratio of Observation to Total Action for "Model with Counters and Dynamics"}
\end{figure}
\twocolumn
\end{document}